\documentclass[sigconf]{acmart}

\usepackage{soul}
\usepackage[some,bottom]{background}

\AtBeginDocument{%
  \providecommand\BibTeX{{%
    \normalfont B\kern-0.5em{\scshape i\kern-0.25em b}\kern-0.8em\TeX}}}

\copyrightyear{2024}
\acmYear{2024}
\setcopyright{rightsretained}
\acmConference[HRI '24 Companion]{Companion of the 2024 ACM/IEEE International Conference on Human-Robot Interaction}{March 11--14, 2024}{Boulder, CO, USA}
\acmBooktitle{Companion of the 2024 ACM/IEEE International Conference on Human-Robot Interaction (HRI '24 Companion), March 11--14, 2024, Boulder, CO, USA}
\acmDOI{10.1145/3610978.3638546}
\acmISBN{979-8-4007-0323-2/24/03}

\begin{document}

\newcommand{\todo}[1]{{\hl{#1}}}
\newcommand{\eg}[0]{\textit{e.g.,}}
\newcommand{\ie}[0]{\textit{i.e.,}}

\title{End-User Development for Human-Robot Interaction}

\author{Laura Stegner}
\authornote{Laura Stegner and David Porfirio are co-chairing this workshop.}
\orcid{0000-0003-4496-0727}
\email{stegner@wisc.edu}
\affiliation{%
  \institution{University of Wisconsin–Madison}
  \country{USA}
}

\author{David Porfirio}
\authornotemark[1]
\orcid{0000-0001-5383-3266}
\email{david.porfirio.ctr@nrl.navy.mil}
\affiliation{%
  \institution{NRC Postdoctoral Research Associate}
  \institution{U.S. Naval Research Laboratory}
  \country{USA}
}

\author{Laura M. Hiatt}
\orcid{0000-0001-5254-2846}
\email{laura.hiatt@nrl.navy.mil}
\affiliation{%
  \institution{U.S. Naval Research Laboratory}
  \country{USA}
}

\author{Séverin Lemaignan}
\orcid{0000-0002-3391-8876}
\email{severin.lemaignan@pal-robotics.com}
\affiliation{%
  \institution{PAL Robotics}
  \country{Spain}
}

\author{Ross Mead}
\orcid{0000-0002-5664-4687}
\email{ross@semio.ai}
\affiliation{%
 \institution{Semio}
 \country{USA}
}

\author{Bilge Mutlu}
\orcid{0000-0002-9456-1495}
\email{bilge@cs.wisc.edu}
\affiliation{%
  \institution{University of Wisconsin–Madison}
  \country{USA}
}

\renewcommand{\shortauthors}{Laura Stegner et al.}

\begin{abstract}
  End-user development (EUD) represents a key step towards making robotics accessible for experts and nonexperts alike. Within academia, researchers investigate novel ways that EUD tools can capture, represent, visualize, analyze, and test developer intent. At the same time, industry researchers increasingly build and ship programming tools that enable customers to interact with their robots. However, despite this growing interest, the role of EUD within HRI is not well defined. EUD struggles to situate itself within a growing array of alternative approaches to application development, such as robot learning and teleoperation. EUD further struggles due to the wide range of individuals who can be considered end users, such as independent third-party application developers, consumers, hobbyists, or even employees of the robot manufacturer. Key questions remain such as how EUD is justified over alternate approaches to application development, which contexts EUD is most suited for, who the target users of an EUD system are, and where interaction between a human and a robot takes place, amongst many other questions. We seek to address these challenges and questions by organizing the first End-User Development for Human-Robot Interaction (EUD4HRI) workshop at the 2024 International Conference of Human-Robot Interaction. The workshop will bring together researchers with a wide range of expertise across academia and industry, spanning perspectives from multiple subfields of robotics, with the primary goal being a consensus of perspectives about the role that EUD must play within human-robot interaction.
\end{abstract}




\maketitle
\SetBgContents{Distribution Statement A. Approved for public release. Distribution unlimited.}
\SetBgScale{1}
\SetBgOpacity{1}
\SetBgVshift{1cm}
\SetBgColor{black}

\section{Introduction}

End-user development (EUD) is an exciting and growing field within human-robot interaction (HRI).
EUD in HRI has roots in enabling nonprogrammers to develop robot applications and has thus been critical to the adoption of consumer robot platforms.
A notable and familiar example is the \textit{Choregraphe} visual programming environment for the Softbank Robotics platforms \cite{pot2009choregraphe}.
Recent successes in academia have explored novel ways to capture user intent (\eg{} through mixed reality \cite{cao2019ghostar} or spoken language \cite{gorostiza2011end}) and provide feedback to users on application quality (\eg{} by verifying ``program correctness'' \cite{porfirio2018authoring}).
Despite these successes, the role of EUD within HRI is still not clear.

There is a pressing need for EUD within HRI to broaden its scope to include related fields.
EUD for HRI has thus far been heavily focused on end-user \textit{programming} \cite{ajaykumar2021survey}, namely the initial creation of programs, whereas EUD encompasses both ``design-before-use'' and ``design-during-use'' \cite{lieberman2006end}, \ie{} the whole life cycle of a program \cite{BARRICELLI2019101}.
As a result, crucial aspects of HRI have been underexplored by EUD, such as the potential  longer-term interactions with robots \cite{irfan2021lifelong}.
Furthermore, within the initial creation of programs, EUD contributions in HRI often provide users with traditional symbolic representations over program logic \cite{ajaykumar2021survey}.
While effective, traditional approaches to programming lead to under-utilization of advances in HRI that may achieve a similar effect, such as learning from demonstration \citep[\eg{}][]{argall2009survey}, natural language communication \citep[\eg{}][]{liu2023grounding}, and personalization \citep[\eg{}][]{lee2012personalization}.
As a result of the focus on ``design-before-use'' and traditional approaches to programming, EUD has yet to fully leverage and provide benefit to the interdisciplinary HRI community. 

In addition to broadening its scope, diverging trends within EUD for HRI create a dire need for consensus.
Crucially, there exists no standard for how EUD research contributions should be evaluated, and little guidance for how contributions should be reported beyond what is presented within past work.
Furthermore, EUD typically falls under systems and technical tracks of HRI publication venues, so the focus of EUD contributions is often on novel standalone programming tools.
Questions thereby remain surrounding open-source expectations of these contributions, how these tools should be maintained and updated, and how EUD contributions could emerge within design and study-focused tracks.
Other relevant questions include how to identify key groups of users that EUD researchers should target and best capture user intent within the myriad of approaches to user interface development.  
For example, many researchers prefer different interface paradigms (\eg{} keyboard-and-mouse tools versus handheld applications) and paradigms for capturing and representing applications (\eg{} block-based versus flow-based programming), yet there exists very little HRI-focused theory on when different paradigms should be used.

The half-day End-User Development for Human-Robot Interaction (EUD4HRI) Workshop at HRI 2024 aims to address the need to broaden the scope of our community, increase interdisciplinary collaborations, and discuss divergent trends in the field. 
We thereby seek to involve a diverse array of perspectives including experts and newcomers to the field from academic and industrial settings. 
We aim to cultivate a body of contributions that stimulate discussion, challenge the current state of the field, and offer novel positions on how EUD can be best situated within HRI.

\section{Outline of the Workshop}

In the EUD4HRI workshop, our goal is to broaden the scope of EUD as it exists within HRI, foster interdisciplinary collaborations with related subfields, assess emerging trends within the field, and find consensus for critical questions that remain within the community.
Our schedule will thereby include discussions, paper presentations, keynote speakers, and panels targeted towards these aims.


\subsection{Format and Activities}

This half-day workshop will be organized into keynote speaker sessions, author presentations, expert panels, and breakout discussions. Panel and discussion topics will focus on opportunities for expanding the scope of EUD within HRI and on the standardization of EUD terminology and use cases in HRI. As part of our goal of including members of underrepresented communities, the format of EUD4HRI will be hybrid.

\subsection{Main Topics of Interest}\label{sec:topics}
Topics of interest include but are not limited to:
(a) novel approaches to application development including underutilized advances in learning and planning, (b) interfaces for robot programming, learning, and personalization, (c) programming languages, representations, and paradigms for HRI, (d) programming libraries and toolkits for HRI, (e) open-source initiatives, (f) standardization, (g) the role of industry and academia within HRI end-user development, and (h) the role of related subfields (\eg{} learning from demonstration, synthesis, personalization, etc.) within end-user development.

\newpage
\subsection{Tentative Schedule (MDT)}
\begin{itemize}
\item[] 09:00 - 09:05: Opening Remarks
\item[] 09:05 - 09:50: Keynote
\item[] 09:50 - 10:35: Author Presentations \#1 
\item[] 10:35 - 11:00: Coffee Break
\item[] 11:00 - 11:30: Author Presentations \#2
\item[] 11:30 - 12:00: Breakout Discussion
\item[] 12:00 - 12:45: Panel 
\item[] 12:45 - 13:00: Closing Remarks

\end{itemize}



\section{Target Audience}
The primary audience of the workshop is a mixture of the following: (1) core members of the EUD for HRI community that can provide insight on the current state of the field and challenges therein; (2) members of related technical fields in HRI from which EUD can draw from; (3) members of fields that apply EUD, such as the child-robot interaction community that often apply EUD interfaces to investigate learning outcomes among children; and (4) members from industry who often utilize EUD advancements within their commercial robots.

Based on past interest in EUD-focused gatherings, we expect to have 15-25 participants. We will adjust the workshop as necessary to include any number of participants, including more or less than the expected amount.

\section{Recruitment of Participants}
Beginning in December, 2023, Calls for participation (CfP) have been distributed within robotics, human-robot interaction, and human-computer interaction mailing lists.
The CfP has also been disseminated through relevant Slack channels frequented by the co-organizers of this workshop and through private email within the organizers' own networks.
The CfP includes our workshop website\footnote{https://sites.google.com/wisc.edu/eud4hri/home} in order to provide prospective participants with as much up-to-date information as possible.

We invite contributions spanning the topics listed in \textit{Main Topics of Interest}, \S\ref{sec:topics}, in the form of short papers of 2-4 pages describing HRI work, blue-sky papers, position papers, \textit{etc}.
PDF submissions are made to EasyChair.\footnote{\href{https://easychair.org}{https://easychair.org}}
We welcome a diverse array of submissions including in-progress work, novel positions on EUD, software libraries and toolkits, and proposed EUD methodology, in addition to traditional research papers and artifact contributions. 
Papers will be peer-reviewed by external members of the HRI and human-computer interaction communities and meta-reviewed by the co-organizers.

Our target number of papers is between 5 and 15, though we welcome and will accommodate as many papers as is appropriate within the time constraints of the workshop.
Based on the number of submissions received, we will invite authors to give either lightning (3-5 minute) or full (8-10 minute) talks with a few minutes reserved for questions for each talk.

\newpage
\section{Documentation and Dissemination}

Paper contributions will be uploaded to the EUD4HRI website prior to the workshop.
At the authors' discretion, paper contributions will also be submitted to ArXiv\footnote{\href{https://arxiv.org}{https://arxiv.org}}.
We will also gauge the interest of participants in the workshop in publishing a review of the workshop outcomes at an appropriate venue. Venues may include HRI'25 Late-Breaking Reports, the AAAI Fall Symposium Series, or a journal.



\section{Acknowledgments}
This workshop is partially supported by the National Science Foundation (NSF) award IIS-1925043 and the NSF Graduate Research Fellowship Program under Grant No. DGE-1747503. 
Additional funding has been received from the EU H2020 SPRING project (grant
agreement No.871245) and from the EU H2020 ACCIO TecnioSpring INDUSTRY (grant agreement
no. 801342, TALBOT project). This work is also supported by the Office of Naval Research and an NRC Research Associateship Program at the U.S. Naval Research Laboratory, administered by the Fellowships Office of the National Academies of Sciences, Engineering, and Medicine. The views and conclusions contained in this document are those of the authors and should not be interpreted as necessarily representing the official policies, either expressed or implied, of the US Navy or the NSF.

\section{Organizers}
\noindent
\textbf{Laura Stegner} is a PhD candidate at the University of Wisconsin--Madison. Her research focus is designing and building end-user development tools that facilitate the personalization of service robots in assisted living settings. Ms. Stegner has previous experience co-organizing workshops centered around the theme of end-user development, including the 2023 AAAI Fall Symposium on \textit{Unifying Representations for Robot Application Development}.

\noindent
\textbf{David Porfirio, PhD} is an NRC Postdoctoral Research Associate at the U.S. Naval Research Laboratory, having received his PhD from the University of Wisconsin--Madison in 2022. His research investigates new ways to design, build, and evaluate software tools for robot application development with a more recent focus on end-user development. Dr. Porfirio has previous experience co-organizing workshops centered around the theme of end-user development, including the \textit{Participatory Design \& End-User Programming} workshop at HRI'22 and the 2023 AAAI Fall Symposium on \textit{Unifying Representations for Robot Application Development}.

\noindent
\textbf{Laura M. Hiatt, PhD} is a research scientist at the U.S. Naval Research Laboratory, where she leads a group that works on a variety of AI research topics. She received her B.S. in Symbolic Systems from Stanford University and her Ph.D. in Computer Science from Carnegie Mellon University. Dr. Hiatt’s work has primarily focused on ways in which humans and robots can work effectively together as teammates. The research involves issues of planning and execution, adjustable autonomy, computational cognitive modeling, and team-based task coordination strategies. Dr. Hiatt currently chairs the AAAI's Education Committee and co-chairs AAAI's committee on US Policy.

\noindent
\textbf{Séverin Lemaignan, PhD} is a senior scientist at PAL Robotics. He received his PhD in joint supervision from LAAS-CNRS and the TU Muenchen, a joint MSc in mechanical engineering from Karlsruhe Institute of Technology and ENSAM ParisTech, and an MSc in Computer Sciences applied to Education from Université Paris. Prior to his position at PAL Robotics, he was an Associate Professor in Social Robotics and AI at the Bristol Robotics Lab, UK. His research spans multiple areas in social robotics including cognition and safe/responsible HRI.

\noindent
\textbf{Ross Mead, PhD} is the Founder and CEO of Semio. He received his PhD and MS in Computer Science from the University of Southern California in 2015, and his BS in Computer Science from Southern Illinois University Edwardsville in 2007.  His research focuses on socially assistive robots, specifically, on the principled design and computational modeling of fundamental social behaviors (e.g., speech, gestures, eye gaze, social spacing, etc.) that serve as building blocks for automated recognition and control in face-to-face human-robot interactions. His research provides the foundations upon which Semio software is being built.

\noindent
\textbf{Bilge Mutlu, PhD} is the Sheldon B. \& Marianne S. Lubar Professor of Computer Science, Psychology (affiliate), and Industrial Engineering (affiliate) at the University of Wisconsin--Madison where he directs the People and Robots Laboratory. His research program focuses on building human-centered methods and principles to enable the design of robotic technologies and their successful integration into the human environment. Dr. Mutlu has an interdisciplinary background that combines design, computer science, and social and cognitive psychology and a PhD in Human-Computer Interaction from Carnegie Mellon University.

\bibliographystyle{ACM-Reference-Format}
\balance
\bibliography{software}

\end{document}